\documentclass{article} % For LaTeX2e
\usepackage{colm2024_conference}
\usepackage{xcolor,colortbl}
\usepackage{amsmath}
\usepackage{textgreek}

\usepackage{microtype}
\usepackage{hyperref}
\usepackage{url}
\usepackage{booktabs}
\usepackage{graphicx}
\usepackage{subfigure}
\usepackage{verbatim}
\usepackage{wrapfig}
\usepackage{lipsum}
\usepackage{dsfont}
\newcommand{\llama}{\textsc{Llama}}
\title{Long Context Alignment with Short Instructions and Synthesized Positions}
\author{
  Wenhao Wu$^\text{\textlambda}$\quad 
  Yizhong Wang$^\text{\textdelta}$\quad 
  Yao Fu$^\text{\textmu}$\quad
  Xiang Yue$^\text{\textpi}$\quad
  Dawei Zhu$^\text{\textlambda}$\quad Sujian Li$^\text{\textlambda}$\\
  $^\text{\textlambda}$Peking University\quad 
  $^\text{\textdelta}$University of Washington\quad $^\text{\textmu}$University of Edinburgh\\
  $^\text{\textpi}$The Ohio State University\\
  \texttt{waynewu@pku.edu.cn}\quad\quad 
  \texttt{lisujian@pku.edu.cn}\quad\quad 
  \\ 
}

\colmfinalcopy
\begin{document}
\maketitle

\begin{abstract}
Effectively handling instructions with extremely long context remains a challenge for Large Language Models (LLMs), typically necessitating high-quality long data and substantial computational resources.
This paper introduces Step-Skipping Alignment (SkipAlign), a new technique designed to enhance the long-context  capabilities of LLMs in the phase of alignment without the need for additional efforts beyond training with original data length.
SkipAlign is developed on the  premise  that  long-range dependencies are fundamental to enhancing  an LLM's capacity of long context.
Departing from merely expanding the length of input samples,  SkipAlign  synthesizes long-range dependencies from the aspect of positions indices.
This is achieved by the strategic insertion of skipped positions within instruction-following samples,  which  utilizes the semantic structure of the data to effectively expand the context. 
Through extensive experiments on base models with a variety of context window sizes, SkipAlign demonstrates its effectiveness across a spectrum of long-context tasks. 
Particularly noteworthy is that with a careful selection of the base model and alignment  datasets, SkipAlign with only 6B parameters achieves it's best performance and  comparable  with strong baselines like GPT-3.5-Turbo-16K on LongBench. The code and SkipAligned models are open-sourced at \url{https://github.com/nightdessert/SkipAlign}
\end{abstract}

\section{Introduction}
The capacity to process and comprehend long contexts is pivotal to large language models (LLMs), empowering them to tackle complex real-world applications involving extremely long context, such as questions answering  or summarizing from multiple-document \citep{caciularu2023peek}, understanding and processing repository-level code \citep{jimenez2023swe}.
Recent advancements have significantly broadened the context window of LLMs, e.g.  achieving a context window of 128K tokens through continuous pretraining \citep{fu2024data}. 
%It is widely accepted that LLMs with an extended context window can deliver enhanced performance on long-context tasks and maintain proficiency on standard short-context benchmarks \citep{xiong2023effective}.

Despite these  advancements on extending context window, the alignment of LLMs to  leverage their long-text capabilities to interpret long and complex instructions remains an underexplored area. 
A primary obstacle is the lack of high-quality, open-source datasets with long instructions, along with the challenges associated with annotating such data.
A promising approach to this challenge involves synthesizing long instructional samples from common short ones.  
However, existing methods have primarily focused on  simply extending the length of instructional samples, neglecting the more critical aspect of effectively building long-range dependency relations.
 For example, methods like LongChat \citep{longchat2023} and Long\llama \citep{tworkowski2024focused} concatenate shorter samples to create longer ones.
Yet, the long-range relations constructed in these strategies are derived from unrelated samples, which  may not effectively simulate the long-range dependencies necessary for tasks involving long context.

To overcome these challenges, this paper introduces a new method called Step-Skipping Alignment (SkipAlign)  which leverages positional indices of short instructions to create samples with meaningful long-range dependency relations. 
Drawing inspiration from transformer's reliance on positional indices, SkipAlign manipulates positional indices to simulate long-range dependencies, enhancing the model's ability to process long contexts without the need for extensive data generation or modifying architecture.
Our technique involves the strategic insertion of skipping steps within the positional indices of instruction-response pairs. 
This strategy is designed to ensure that the relative distances of synthesized indices  are uniformly distributed across an extended range of lengths, while maintaining their continuity  as much as possible.
Leveraging the rich long-range dependencies within the synthesized positions, LLMs are better equipped to learn how to process long  instructions during the alignment phase.

Our evaluation of SkipAlign involved base models with varying context window sizes, including a \llama-2 model featuring a 4096-token window and a Yi-6B-200K model with an  200K-token window.
On  LongBench benchmark, SkipAlign activates long-context capabilities more effectively than conventional instruction  finetuning and recent packing based methods.
A SkipAlign model with 6 billion parameters, when integrated with high-quality base models and instruction datasets, matches the performance of GPT-3.5-Turbo-16k on the LongBench. 
Moreover, in the Needle-in-a-Haystack test, SkipAlign  demonstrates its superior performance in  extending the context window size and  highlights the critical importance of long-range dependencies in samples, rather than merely extending the sequence lengths.
In summary, the advantages of SkipAlign are as follows:
(1) \textbf{Enhanced Long Context Capabilities}: SkipAlign improves models' long context capabilities  by simulating long-range dependencies, which is essential for effective long context alignment.
(2) \textbf{Computational Efficiency}: SkipAlign avoids the need for additional longer data for training or modifying the architecture of a LLM, making it a computationally efficient solution.
(3) \textbf{Extended Context Window}: SkipAlign additionally helps LLM with small context window to handle inputs beyond their original context window.
\begin{figure}
    \includegraphics[width=1\linewidth]{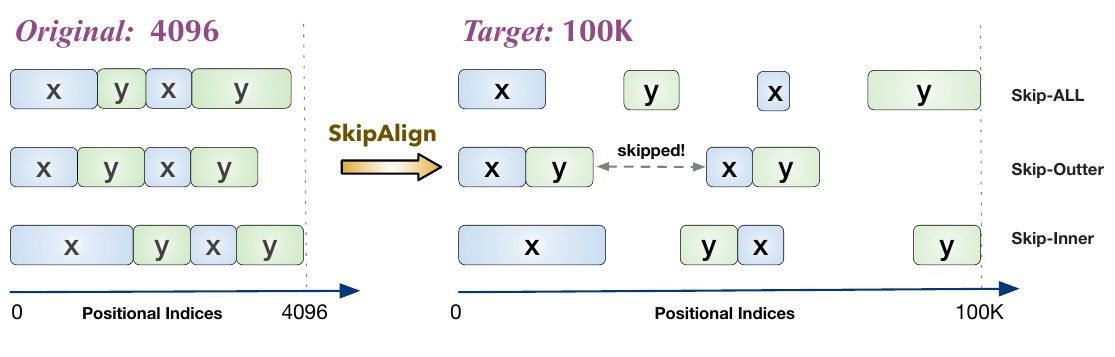}
    \caption{SkipAlign modifies positional indices in instruction-following samples to simulate long-range dependency relations. The provided example   showcases how SkipAlign takes three distinct samples, each initially positioned within a 4096-token, and independently applies three separate strategies to stretch their lengths to an impressive 100K tokens. }
  \label{fig:method_illus}
\end{figure}
\section{Related Work}
\paragraph{Long Context Scaling} 
The goal of long context scaling is to empower current LLMs them with the ability to cope with long context tasks. 
This process involves two key steps: context window extension and instruction finetuning \citep{xiong2023effective}.
The majority of existing research has concentrated on the former, exploring techniques such as manipulating positional embeddings \citep{chen2023extending, ntk, jin2024llm}, innovating model architecture  \citep{mohtashami2023landmark, yang2023gated, tworkowski2024focused}, and continue pretraining \citep{chen2023longlora}.
In contrast, this study delves into the latter step, focusing on long context instruction finetuning. To the best of our knowledge, previous research has approached this stage by generating additional long-input data \citep{bai2024longalign}. 
Our method, however, relies solely on the available short instruction data.
\paragraph{Long Context Evaluation} Initial studies have predominantly evaluated LLMs based on their ability to maintain perplexity over extended context \citep{chen2023extending, peng2023yarn}.
 However, recent findings have revealed that  perplexity  alone is insufficient to reflect the long context capabilities of language models \citep{fu2024data}. 
As a result, two alternative evaluation methods have emerged.
One approach involves comprehensive evaluation methods,  such as LongBench 
 \citep{bai2023LongBench} and L-Eval 
 \citep{an2023eval},
which assess long context capabilities through various downstream tasks, including question answering (QA) and text summarization.
The other approach, represented by Needle-in-a-Haystack test\footnote{https://github.com/gkamradt/LLMTest\_NeedleInAHaystack.},
 applies synthetic tasks to pressure test specific types of long context capabilities at any given position.
In addition to assessing long context capabilities, it is crucial to evaluate a model's proficiency in managing short texts effectively \citep{xiong2023effective}. 
In this paper, we conduct a comprehensive evaluation by employing both types of long context evaluation methods, while also reporting on the performance of short text tasks.
\paragraph{Skip Position Training}
The concept of skip position training has been previously utilized for context window expansion.
RandPos \citep{ruoss-etal-2023-randomized} randomly selects and projects an ordered subset of position indices to accommodate longer contexts.
Subsequently, PoSE \citep{zhu2023pose} refined this technique by dividing long inputs into segments and randomly shifting their position indices. 
The primary objective of these methods is to enhance memory efficiency during the training of extremely long sequences. Our approach, on the other hand, aims to stimulate long-range dependencies in long instruction-following data and  utilizing their inherent structure.

\section{Methodology}
\subsection{Preliminary}
Before introducing SkipAlign, we first introduce the background knowledge and the important baselines of our method.
\paragraph{Instruction Tuning} \label{instruct_tuning}
Pretrained models are often finetuned  with instruction-following samples for alignment to learn to follow instructions. 
These samples are structured as  instruction-response pairs, arranged in continuous sequences \citep{wei2022finetuned}. 
These sequences are structured as formal instruction-response pairs. 
To formalize, let $m=(x_1, y_1, \dots, x_i, y_i)$ denote a sequence comprising $i$ turns of such pairs.
We train auto-regressive language models using the following objective function:
\begin{equation}
\mathcal{L} = -\sum_m\log\sum_{y_j} p(y_j | (x_1, y_1, \dots, x_j)),\label{eq:loss}
\end{equation}
In this dialogue-formatted sample, the model is tasked with predicting each response $y_j$  conditioned on its preceding instruction  $x_j$ and the sequence of prior pairs.
This conventional approach to instruction tuning is termed \textit{Normal-SFT} throughout the remainder of this paper.

\paragraph{Packed-SFT}
It is crucial to highlight that the majority of existing datasets used for instruction tuning are characterized by short instructions.
To address this limitation, a straightforward method proposed in LongChat \citep{longchat2023} involves concatenating multiple short, unrelated instruction-following samples into a single sequence of $k$ tokens in length.
We refer this baseline method as \textit{PackedSFT-k} throughout the remainder of this paper.

\paragraph{Position Indices}  Transformer-based language models utilize positional information to complement the input tokens, and this information is encoded through positional indices \citep{vaswani2017attention}. 
While a variety of positional embedding techniques have been proposed, they universally rely on positional indices to precisely convey the positional information of tokens \citep{raffel2020exploring, su2024roformer}.
 By default, positional indices are sequentially assigned as $(0, 1, \dots, |m|-1)$, with $|m|$ representing the length of the input sequence.
In this study, we concentrate on the recent popular  relative positional embedding approach,  with a particular emphasis on the ROPE \citep{su2024roformer}.
This method characterizes the positional relationship between two tokens at indices
$i$ and $j$ by their relative distance, denoted as $|i - j|$.

\subsection{SkipAlign}
In this section, we provide an in-depth explanation of our proposed method, SkipAlign.
To generate a target response within an instruction-following sample, the essential information relied upon is scattered across its corresponding instruction and the sequence of preceding dialogue turns, as elaborated in Section \ref{instruct_tuning}. 
SkipAlign operates on the core assumption that expanding the relative distance of such semantic structure to encompass a longer scale is essential for unlocking the long-context capabilities of language models.
SkipAlign accomplishes this via strategically modifying positional indices. 
By selectively skipping over certain positional indices in a instruction-following sample, we are able to extend the relative distance of semantic dependencies, creating long-range dependency relations.%, thereby enhancing the model's ability to process and comprehend long-form contexts. 

\paragraph{Skipping Positions via Shifting} 
Our aim is to expand relative distances of  semantic dependency  in  an instruction dataset, surpassing the its maximum sample length 
$l$ to reach an extended maximum length 
$L$, where 
$L$ is significantly greater than $l$.
This is achieved by reassigning positional indices,  spreading the original positions from the interval  $[0, l]$ to the extended interval $[0, L]$.
We treat an instruction or response as a basic unit and  shift  all of their positional indices simultaneously.
Formally, given an $i$ turn sample $m$,  let $P(m) = (\boldsymbol{c}_1, \boldsymbol{c}_2, \dots, \boldsymbol{c}_{2i-1}, \boldsymbol{c}_{2i} )$ represent its original positional indices which is concatenated by  the positional indices of each block in a instruction-response pair.
In $P(m)$,  odd  and  even numbered subscript separately correspond to  instructions and responses.
 We create larger relative positions by shifting each positional block to the right by a bias vector $\boldsymbol{u} = (u_{1}, u_{2}, \dots u_{2i})$, where each constant $u \in \boldsymbol{u}$  is a constant bias for the shift.
 By shifting different block by a various scale, we can create skipping positions between them.
The reassigned positional indices of $m$ are now given by:
\begin{equation}
P_u(m) = P(m) + \boldsymbol{u} = (\boldsymbol{c}_1 + u_1, \boldsymbol{c}_2 + u_2, \dots, \boldsymbol{c}_{2i} + u_{2i} ).
\end{equation}
Because the basic requirement for valid position indices is incrementality, which requires the minimum shifting bias $u_i$ is set  to accumulated shifting bias of previous 
 tokens $u^{a}_i = \sum_{j<i}u_j$.
We introduce a skipped step denote as $s_i$,  such that  $u_i = u^a_{i-1} + s_i$.
A $s_i$ of zero means no skip occurs between  $\boldsymbol{c}_{i}$ and  its precedent $\boldsymbol{c}_{i-1}$.
A positive $s_i$ introduces a skip of  $s_i$ positional indices between these two positions.
To achieve a uniform distribution of relative distances within $[0, L]$ after shifting, we sample $s_i$ from a uniform distribution:
\begin{equation}
    s_i \sim \mathcal{U}\{1, L-|m|-u^a_{i-1}\},
\end{equation}
 where $L-|m|-u^a_i$ represents the maximum allowable skip length, taking into account the sample length $|m|$ and the already skipped positions $u^a_{i-1}$.
 The remaining critical task is to devise a skipping strategy for determining when to set $s_i>0$ to introduce skipping steps.
\paragraph{Skipping Strategy}
We investigate three distinct skipping strategies, to study the contributions of various semantic dependencies on the model's long context capability. These strategies apply skipped distances selectively to particular structures within the sample:
\begin{enumerate}
    \item \textbf{Skip-All}:  This strategy applies skipping across all roles within a sample, without any selection.
    \item \textbf{Skip-Inner}: This strategy  adds skipping steps exclusively within pairs, i.e., between an instruction and its response. Concisely, such strategy only adds $s_i$ when $c_i$ is from a response.
    \item  \textbf{Skip-Otter}: This strategy introduces skipping steps only between separate dialogue turns. Concisely, such strategy only adds $s_i$ when $c_i$ is from a instruction.
\end{enumerate}
A straight forward illustration of how these strategies on positional indices is presented in Figure \ref{fig:method_illus}.
We  use an indicator function \textbf{DO\_SKIP}() to determine if $c_i$ meets the criteria for adding skipping step. 
The function returns 1 if the conditions are met, and 0 otherwise. 
Furthermore, to control the number of synthesized positions, we sub-sample  $p\%$ of valid position to add  skipping steps.
 The overall rule are summarized as  followings:
\begin{equation}
    u_i = 
    \begin{cases}
        u^a_{i-1} +\mathds{1}(\epsilon_i \leq p) *s_i & i > 0  \text{and}   DO\_SKIP(c_i) \\
        0 & i=0
    \end{cases},
\end{equation}
where $\epsilon_i$ is uniformly sampled from $[0, 1]$ and determined by the indicator function $\mathds{1}(\cdot)$, which decides whether to add the skipped distance $s_i$. 
We apply \textbf{Skip-Outer} as our default strategy as it achieve a better performance in both long context  and short context  capability by  ablation studies (\ref{exp:shot_task}). 
\begin{wrapfigure}{r}{0.5\textwidth}
\vspace{-0.1cm}
\centering
\includegraphics[width=0.5\textwidth]{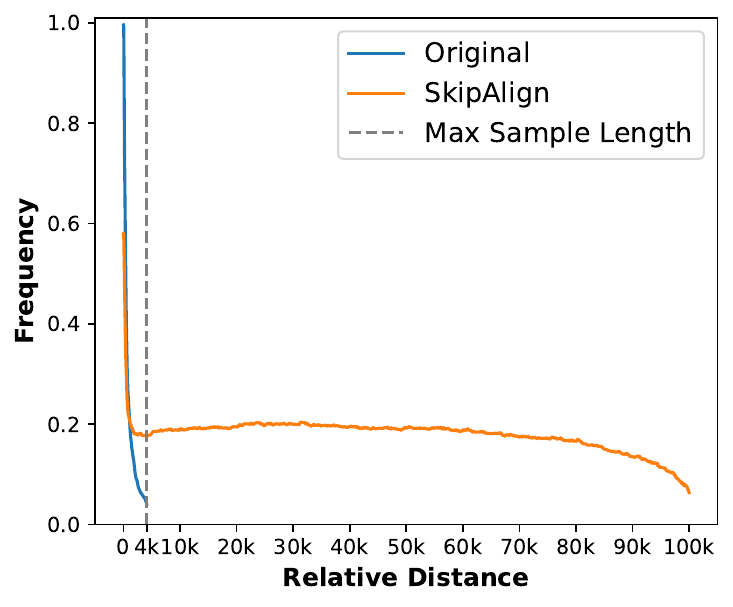}
\caption{The frequency of relative distance in the T\"ulu V2 dataset. Comparing with the original distribution, SkipAlign redistribute a small subset of samples into a longer context. \vspace{-0.3cm}}
\label{fig:pos_dis}
\end{wrapfigure}
\paragraph{Frequency of Relative Distances}
Distribution of relative distances within a dataset is the key to understand the impact of the SkipAlign. 
This section provides a statistical analysis of the frequency of relative distances at the dataset level.
We begin by explaining the methodology to quantify the range of relative distances present in an individual sample.
In the most straightforward scenario,  a single-turn dialogue $(x_1, y_1)$ with a length of   $l$, the set of possible relative distances  for generating $y_i$ is $\{0, 1, \dots, |l|-1\}$.
However, if a skipped step $s_i$ is inserted between $x_1$ and $y_1$, the minimum distance between them is now  $s_i$, the revised range of relative distances is $\{s_i, s_i + 1 + \dots, s_i + |l|-1 \}$, which  expands the relative distance of such dependency.
For more complex cases involving multiple turns, we consider the union of the relative distance sets for generating responses in each turn.

Following the aforementioned mehotd, we calculate the frequency of relative positions in dataset-level. %
%To exemplify the effects of SkipAlign, we analyze the frequency of relative positions in a sample  from the T\"ulu V2 dataset \citep{ivison2023camels}
As depicted in Figure \ref{fig:pos_dis}, T\"ulu V2 dataset's initial relative distances are confined to the interval $[0, 4096]$.
After SkipAlign the distribution is extended to $[0, 100K]$, with the extended range from 4096 to 100K  nearly uniform.
This observation suggests that the SkipAlign extends the positional indices of a $p\%$ of the dataset, making them to  evenly distributed to relative distances across the expanded interval.

\begin{comment}
    To simplify the effort of analzationand also because most of the samples in instruct dataset only contain 1 turn of dialogue, we analyze a dataset only contains of one pair of  $ (x_1, y_1)$.
Denote this pair has a length of $l$, it is easy to infer that the relative distance appeared in generating responses $y_1$ are in the range of $\{0, 1, \dots, |l|\}$.
In the case of fake positions introduced above, if there are a skipped distance between $x_1$ and $y_1$, the appeared relative distances will be in $\{s_i, s_i + 1 + \dots, s_i + |l| \}$.
Considering in the whole dataset with maximun sample length of $l$,  the  frequency  of  relative distance $j$ will be $f_j$.
Denote this pair has a length of $l$, it is easy to infer that the relative distance appeared in generating responses $y_1$ are in the range of $\{0, 1, \dots, |l|\}$.
In the case of fake positions introduced above, if there are a skipped distance between $x_1$ and $y_1$, the appeared relative distances will be in $\{s_i, s_i + 1 + \dots, s_i + |l| \}$.
Considering in the whole dataset with maximum sample length of $l$,  the  frequency  of  relative distance $j$ will be $f_j$.
\end{comment}

\section{Experimental Setup}
\paragraph{Training Data}
Our experiments leverage the T\"ulu V2 \footnote{https://huggingface.co/datasets/allenai/T\"ulu-v2-sft-mixture} dataset, which is a high-quality data mixture consisting of manually annotated and GPT-generated conversational data. 
This dataset provides a rich and diverse source for model training.
Following their settings, we  truncate input samples  to 4096 tokens. 
For the SkipAlign, we introduce additional positional indices during pre-processing. 
The parameters for the SkipAlign are as follows: the maximum extend length 
$L$  is set to 100K, the sub-sampling ratio $p$ is 0.5, and the default skipping strategy is Skip-Outter.

\paragraph{Training Settings} 
In response to the recent progress in extending the context window, our study investigates  the influence of these models on the alignment of long contexts. 
 We conduct our SFT experiments using two base models with varying context window sizes:
1. The \llama-2 model \citep{touvron2023llama}, which has a context window of 4094 tokens, serves as our baseline for comparison.
2. The Yi-6B-200K model \footnote{https://huggingface.co/01-ai/Yi-6B-200K}, which significantly extends the Yi-6B model's context window to an impressive 200K tokens through continuous pre-training \citep{ai2024yi}.
For models based on \llama-2, we employ the Neural Tangent Kernel (NTK) \citep{ntk} to extend positional embeddings to the maximum training or inference length prior to training. In contrast, for Yi-6B-200K models, additional positional extension is unnecessary as the model's inherent maximum embedding length is already 200K.
For the hyper-paramter settings for training, please refer to Appendix \ref{sec:paramter}.

\paragraph{Evaluation} 
The evaluation of our models' performance with long contexts is conducted using LongBench \citep{bai2023LongBench}, a comprehensive benchmark suite that encompasses 16 distinct datasets spread across 6 different task categories. These datasets are designed to assess models with input lengths varying from 4K to 20K tokens.
In the course of our experiments, we observed significant instability in the performance of synthetic tasks within LongBench when tested across multiple models and even at different checkpoints within the same model. 
This variability prompted us to exclude synthetic tasks and any Chinese-language datasets from our evaluation to ensure a more reliable and focused assessment.
 We set the maximum testing length to 16K tokens.

\section{Results}

\begin{table*}[t]
\centering
 \setlength\tabcolsep{5pt}
\begin{tabular}{lccccccc}
\toprule
\textbf{Model}&\textbf{Avg.}&S-Doc QA&M-Doc QA&Summ&Few-shot&Code\\
\midrule
GPT-3.5-Turbo-16k & 44.6	&39.7	&38.7	&26.5	&67.0	&54.2 \\
\midrule
\textbf{\llama-2-7B Based Models}\\
\llama-2-7B-chat-4k&35.2&24.9&22.5&25.0&60.0&48.1 \\
SEext-\llama-2-7B-chat-16k&	38.7&	27.3&	26.2&	24.8&	64.2&	57.5\\
LongChat1.5-7B-32k&	36.9&	28.7&	20.6&	26.6&	60.0&		54.2\\
\llama-2-7B-NTK32k&31.7&16.2&	7.3&	15.4&	66.7	&	\textbf{63.4}\\
\quad + Normal-SFT&41.5&31.3&	32.7&	26.0&	65.3& 57.4\\
\quad + PackedSFT-16k&	42.6&31.6&	32.8&	26.2&	67.9&		60.5\\

\quad + PackedSFT-32k&41.6&30.0&32.2&	26.2&	67.3&58.0\\
 \quad + PackedSFT-50k&	43.6&	36.0&	\textbf{37.0}	&\textbf{27.7}&	63.8&		58.5\\
 \rowcolor{yellow}
 \quad + SkipAlign &\textbf{44.1	}&\textbf{38.6}&	33.8&	26.1&	\textbf{67.6}&59.6\\
\midrule
\textbf{Yi-6B-200K Based Models}\\

%Yi-6B&36.9&22.5&26.6&21.0&62.8&58.6\\
%Yi-6B-chat-NTK&37.0&19.1	&25.4	&23.0	&65.1		&60.1\\
Yi-6B-200K&39.1&25.1&	33.8&	25.6&	56.6&		\textbf{62.0}\\
\quad + Normal-SFT &43.7&37.0&35.0&26.8&	65.8&	59.0\\
\quad + PackedSFT-16k&44.1&33.1&38.2&\textbf{27.4}&\textbf{67.4}&		59.7\\

\rowcolor{yellow}
\quad + SkipAlign 				&\textbf{45.3}&\textbf{40.3}	&\textbf{38.7}	&26.1	&66.3	&60.0 \\

\bottomrule
\end{tabular}
\caption{Results on LongBench, we report the average performance on all datasets and each sub tasks of various long context alignment settings.}
\label{exp:LongBench}
\end{table*}

\subsection{Results on LongBench}
We present the results of our comprehensive experiments on LongBench in Table \ref{exp:LongBench}.

\paragraph{SkipAlign further benefits long context capability}
The results presented in the second and third blocks of Table \ref{exp:LongBench} highlight the consistent advantage of SkipAlign over Normal-SFT and  Packed-SFT on average scores.
This is particularly evident  when comparing with Noraml-SFT, where SkipAlign almost demonstrates its superiority in every subtasks.
Utilizing the Yi-6B-200K model, SkipAlign outperforms GPT-3.5-Turbo-16k in the overall average performance on LongBench.

\paragraph{Task-level Analysis} 
After alignment, there is a noticeable enhancement in performance across all sub-tasks, with the exception of a slight decline in the coding subtask.
This is largely attributed to the fact that the coding tasks in LongBench predominantly involve continuous code generation, a type of task that aligns more closely with the pretraining.
Models need to pay ``alignment tax'' for this task.
In task-level comparisions, the improvements brought by SkipAlign, in descending order, are single-document QA, multi-document QA, few-shot learning, and lastly, summarization.
The driving force behind these improvements is SkipAlign's proficiency in simulating long-term dependencies.
Conversely, the gains observed in summarization tasks were more modest. This can be explained by the complex nature of information aggregation inherent in summarization. The task requires identifying  salient information that is evenly dispersed throughout a long context. 
Constructing this type of long-term structure is challenging for current skipping strategies, which are constrained by the given short data and the necessity to maintain consistency of their positional indices.

\paragraph{Quality of base model and alignment dataset is important to the long context capability} 
Our investigation has revealed key insights into how the quality of base models and alignmnt datasets significantly influence a language model's ability to handle long contexts.
Notably, when using the same SFT dataset, Noraml-SFT, PackedSFT-16K, and SkipAlign consistently show more improvements when they are based on the Yi-6B-200K model rather than the \llama-7B model. 
Moreover, despite employing a similar packing strategy and training sequence length, the PackedSFT-32K model, trained with the Tülü V2 dataset, outperforms the LongChat1.5-7B-32k model, which was trained using ShareGPT, by a notable 4.7 points.
This observation underscores the importance of both a high-quality alignment dataset and s base model with inherent strong long context capabilities in achieving superior overall performance.

\subsection{Testing with Needle-in-a-Haystack}
\begin{figure}[t]
  \centering
  \subfigure[\llama-2-7B-NTK-50K]{
    \includegraphics[width=0.48\textwidth]{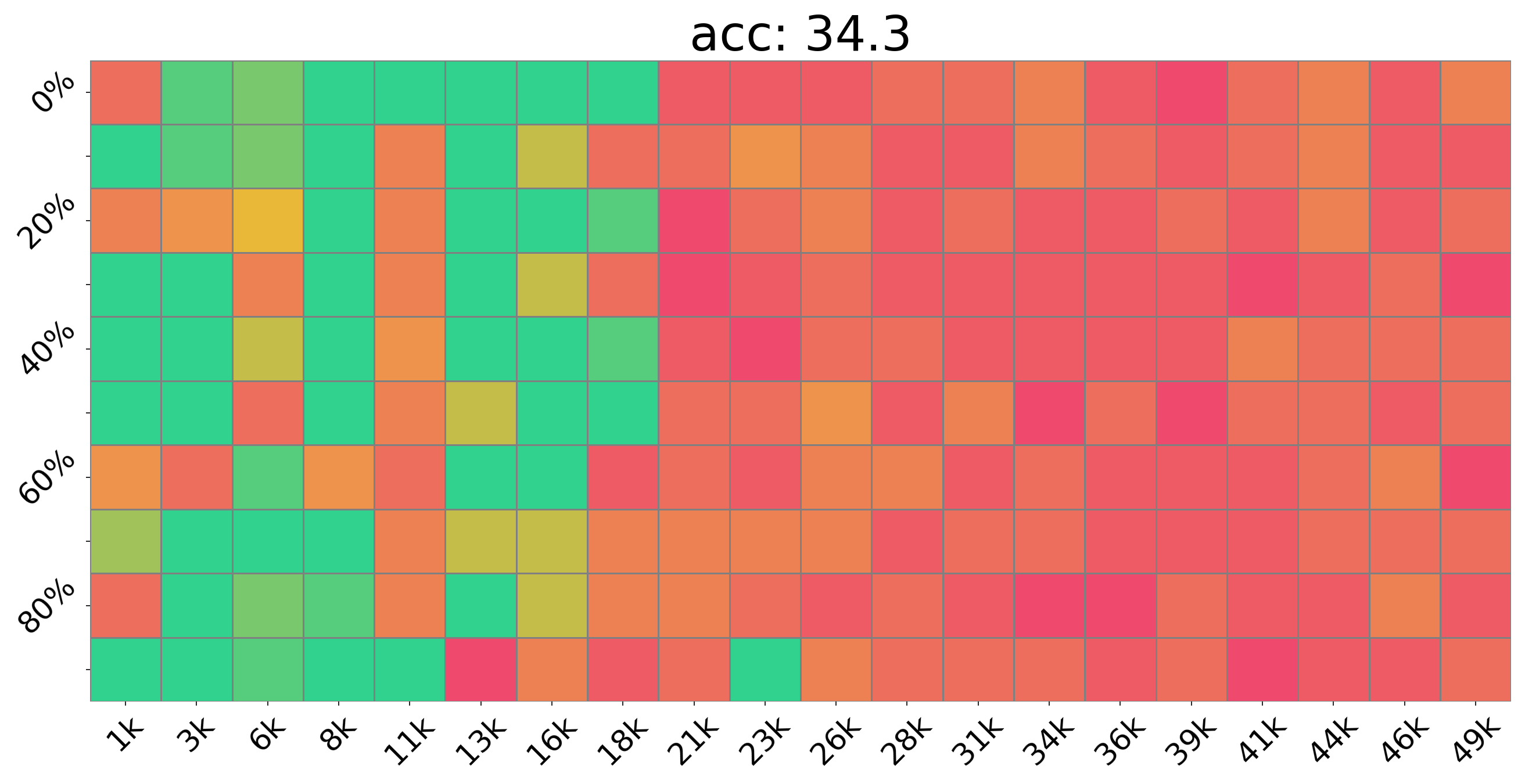}
    \label{fig:ntk-50k}
  }
  \subfigure[Normal-SFT-NTK-50K]{
    \includegraphics[width=0.48\textwidth]{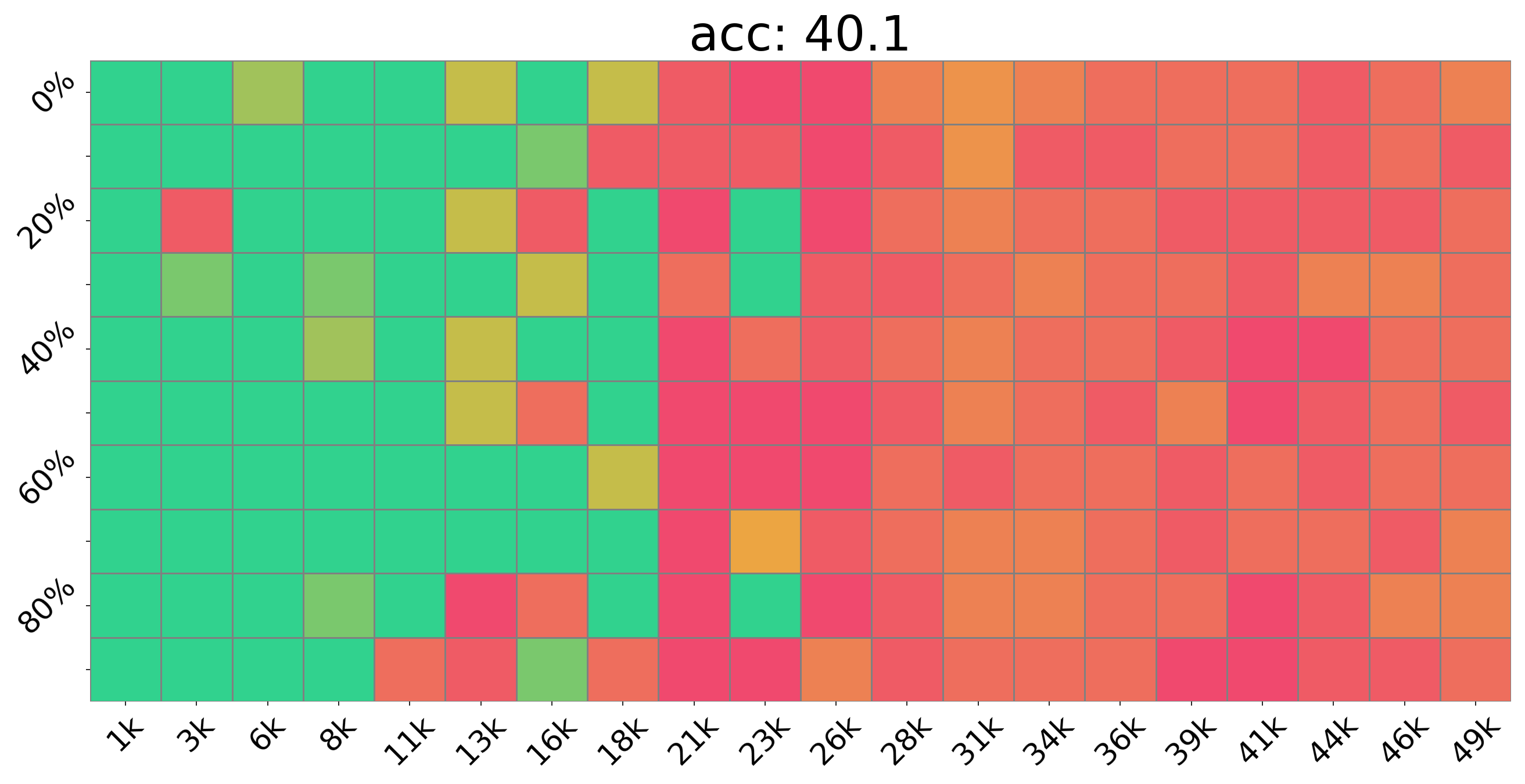}
    \label{fig:noraml_sft}
  }
    \subfigure[PackedSFT-50K]{
    \includegraphics[width=0.48\textwidth]{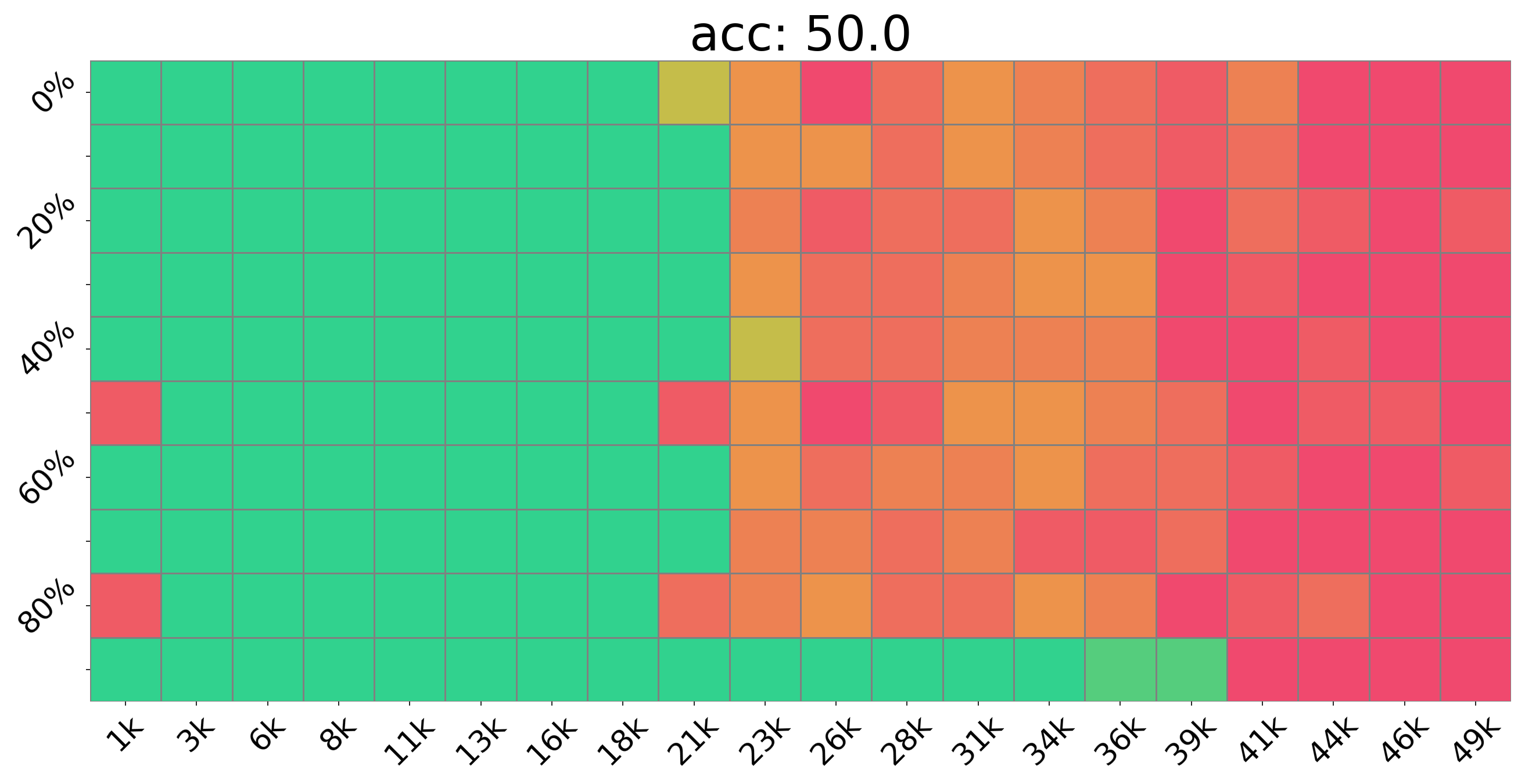}
    \label{fig:llama_packed16k}
    
  }
  \subfigure[SkipAlign]{
    \includegraphics[width=0.48\textwidth]{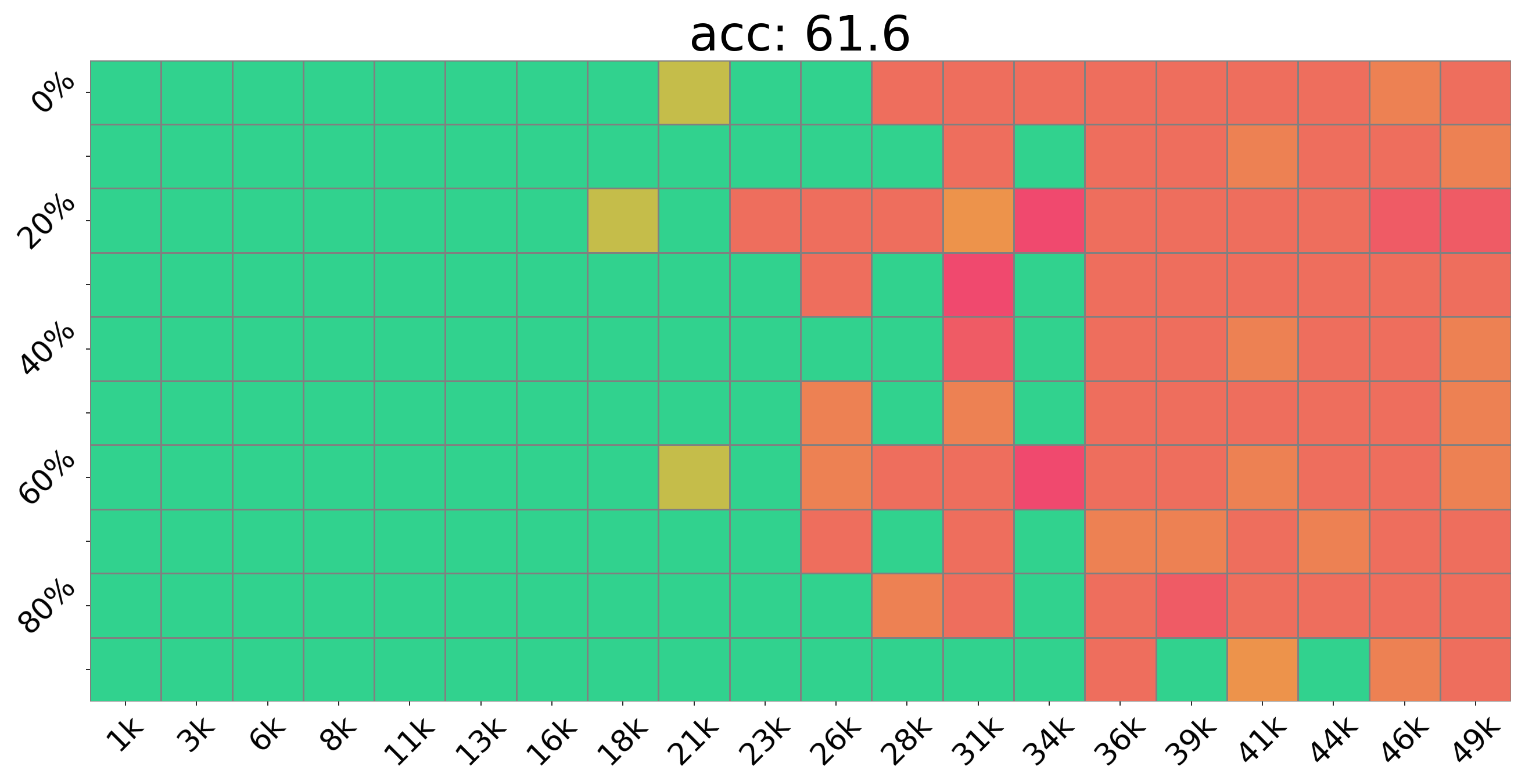}
    \label{fig:llama_skip}
  }
  \caption{Needle in the Haystack test for \llama-2-7B  based models: \llama-2-7B-NTK-50K denotes the straightforward expansion of \llama-2-7B using NTK to accommodate 50K tokens without further tuning. Normal-SFT-NTK-50K represents the adaptation of a standard fine-tuned model for this extended context. PackedSFT-50K indicates the fine-tuning process using samples artificially extended to 50K tokens for training.}
  \label{fig:needle}
\end{figure}

\paragraph{Settings}
To gain a clearer insight into the enhancement of long context capabilities by SFT and our proposed SkipAlign, we conduct a Needle-in-a-Haystack test. 
This test evaluates a model's ability to retrieve information from any position within the context, as depicted in Figure~\ref{fig:needle}. 
We use a color scale ranging from deep red, indicating a 100\% successful recall, to green, representing a 0\% complete failure.
Given that the Yi-6B-200K model has already achieved near-perfect performance in this test, we focus our evaluation on \llama-2-7B based models.

\paragraph{SkipAlign is better at extending context window} 
Directly applying NTK for inference, as shown in Figure \ref{fig:ntk-50k}, yields suboptimal results. While initial fine-tuning followed by NTK, as depicted in Figure \ref{fig:noraml_sft}, slightly expands the context window beyond the initial 4096 token limit.
Conversely, fine-tuning with packed samples to accommodate a 50K token context, as illustrated in Figure \ref{fig:llama_packed16k}, manages to extend the successful retrieval window to around 20K tokens, achieving an average accuracy score of 50. 
However, SkipAlign (Figure \ref{fig:llama_skip}), which does not rely on samples exceeding 4096 tokens, not only extends the retrieval window to a  extent of 28K but also significantly improves the average accuracy score to 61.6. 
This outcome demonstrates SkipAlign's superior ability to enhance the context window without the need for excessively long input samples.

\paragraph{Long-term dependency are more important than sample's length}
A detailed comparison between PackedSFT-50K and SkipAlign reveals the critical role of long-term dependencies. 
With PackedSFT-50K, the input sample size is uniformly concatenated to 50K tokens, ensuring that each sample reaches this length.
In contrast, SkipAlign employs a strategic approach to enhance long-term dependencies without necessitating the creation of actual long samples.
From the perspective of relative distance, although PackedSFT-50K samples are longer, the effective dependency relationships they capture are confined within a 4096 token relative distance. 
SkipAlign, on the other hand, explicitly extends these relationships to a much broader range.
This under-scoring the notion that the effective
long-term dependencies is a more critical factor  than the mere length of the input sequences.

\subsection{Ablation Study on  short text capability and  on skipping strategy}
\paragraph{Evaluation Settings }
In addition to the long context evaluation previously discussed, we conducted further tests to determine the influence of various SFT  configurations on a model's fundamental short text processing capabilities. 
Following the evaluation settings in \citet{wang2023far}, we validate on 6  datasets: Massive Multitask Language Understanding dataset (MMLU \cite{hendrycks2020measuring}) for measuring models’ factual knowledge, and Big-Bench-Hard (BBH \citep{Suzgun2022ChallengingBT})  to evaluate models’  reasoning capabilities, TyDiQA to evaluate models’ multilingual capabilities \cite{clark2020tydi}, and Codex-Eval which based on HumanEval dataset \cite{Chen2021EvaluatingLL} to evaluate coding capabilities.
\begin{table}[t]
\centering
 \setlength\tabcolsep{5pt}
\begin{tabular}{lc|ccccc}
 \toprule
 Model&LongBench&MMLU&BBH&TydiQA&Codex-Eval \\
 \midrule
 Yi-6B-200K&39.1&64.2&43.0&16.2&19.9\\
 \quad +Normal-SFT&43.7&60.5&	44.6&	32.6&	30.4\\
 \quad +Skip-All&45.1&	59.6&	38.7&	31.7&	26.9\\
\quad +Skip-Inner&42.4&59.5&	41.5&	31.0&	29.3\\
\quad +Skip-Outter (default)&45.3&61.1&	42.6&	30.3&	28.5\\
 
 \bottomrule
\end{tabular}
\caption{Results on both long and short tasks. }
\label{exp:shot_task}
\end{table}

\paragraph{Trade-offs in SkipAlign's Performance} 
Since SkipAlign samples a subset of the data to synthesize long range dependency, thereby reallocating computational resources that would have been directed towards short-text processing to optimize the handling of longer sequences.
As illustrated in Table \ref{exp:shot_task}, since the overall content of the data remaining unchanged, SkipAlign doesn't affect the learning of factual knowledge.
In fact, it shows a  improvement of 1.5 points on the MMLU metric when compared to Normal-SFT.
For the performance on BBH (Resoning), TydiQA (multilingual) and Codex-Eval (Coding), SkipAlign witness a 1-2 point decrease, which could potentially be attributed   to the selective nature of SkipAlign.
In summary, SkipAlign strategically shifts some of the short-text capabilities of Normal-SFT to enhance its long-context performance. 

\paragraph{Integrity of dialogue structure is crucial for SkipAlign}
The integrity of the dialogue structure, specifically the consistency between instructions and responses, is crucial for sustaining performance across both long and short text tasks. 
When skipping steps are applied within an instruction-response pair (Skip-Inner), it negatively impacts the model's performance, regardless of the text length.
Interestingly, the Skip-All strategy, which applies skipping without any constraints, achieves a performance  that lies between the extremes of Skip-Inner and Skip-Outter. 
This observation highlights the significance of maintaining the integrity of the dialogue structure.
\subsection{Analysis on Hyper-parameter}

\begin{wrapfigure}{R}{0.5\textwidth}
\vspace{-1cm}
\centering
\includegraphics[width=0.55\textwidth]{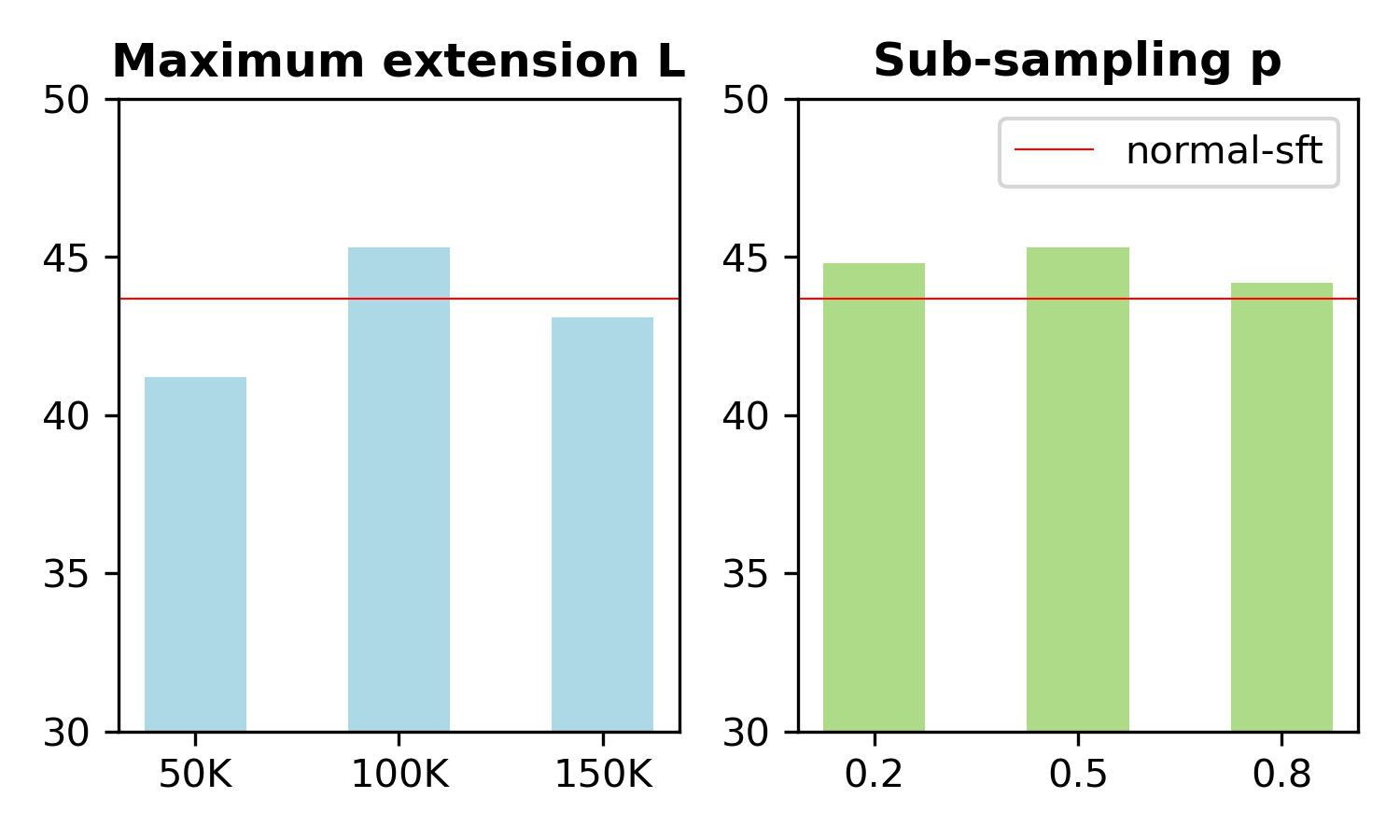}
\caption{Average score on LongBench for SkipAlign aross various maximum extension length $L$ and sub-sampling ratio p $p$.\vspace{-0.4cm}}
\label{fig:parameters}
\end{wrapfigure}

We conducted a thorough analysis of how various hyper-parameters influence  SkipAlign's performance. 
\paragraph{$L$ effects overall performance most, with 100K being  the optimal setting}
Figure \ref{fig:parameters} demonstrates that, in comparison to  $p$, severely affect the overall performance of SkipAlign.
Among the evaluated lengths, $L$ set to 100K stands out as the most effective, consistently delivering superior results to both the Normal-SFT and the lengths of 50K and 150K.
It is noteworthy that the average testing length on  LongBench dataset is below 50k, suggesting  that utilizing a $L$ that significantly larger $l$, such as 100K or 150K, can lead to better performance after alignment.
\paragraph{A moderate setting of $p$ yields optimal performance} 
With $p$ across 0.2, 0.5, and 0.8, SkipAlign consistently outperforms Normal-SFT and achieves peak performance at a probability of 0.5. 
 This peak indicates that a moderate value of $p$ enables SkipAlign to optimize its performance effectively.
\section{Conclusion and Future Research}
In this study, we introduce SkipAlign, a new method designed to perform long context alignment only with short instruction datasets. 
This technique employs a simple yet effective strategy of manipulating position indices within instruction-following samples, thereby facilitating the creation of high-quality long dependency relations. 
Our extensive experimental evaluation across a variety of long-context tasks demonstrates have consistently shown that SkipAlign surpasses conventional instruction finetuning and other long-context synthesis methods in performance.
A key insight from our Needle in the Haystack experiment is that the length of the training samples is secondary to the establishment of effective long-term dependency relations, which are crucial for mastering long-context capabilities.

Looking ahead, there are two principal directions for the  future  studying  of SkipAlign:
The first direction involves a deeper investigation into long-context alignment. Given that SkipAlign currently synthesizes features with only short samples, future work could examine its performance when integrated with actual, annotated long-context examples. This integration could potentially enhance the model's ability to generalize  to even longer contexts.

The second direction focuses on extending the context window through continued pretraining. 
With the recent advancements in causal continuous pretraining, such as models reaching 120K tokens trained on 80G GPUs, there is an opportunity to explore whether SkipAlign's skip-position training can elevate these pretraining efforts to accommodate even longer contextual lengths, for example 1M context length.
This exploration could lead to LLMs with unprecedented context windows, significantly expanding their applicability in complex long tasks.
\bibliography{colm2024_conference}
\bibliographystyle{colm2024_conference}

\appendix
\section{Hyper-parameters for Training}\label{sec:paramter}
All models are trained for two epochs with a learning rate of 1e-5, without weight decay, and using a linear learning rate decay and linear warmup for 3\% of the total training steps. Training is conducted on an 8-GPU setup with NVIDIA A100 GPUs, utilizing the DeepSpeed library \cite{aminabadi2022deepspeed} and the ZeRO optimizer \citet{Vaswani+2017} for efficient and stable training.

\end{document}